\documentclass[letterpaper, 10 pt, conference]{ieeeconf}
\IEEEoverridecommandlockouts
\title{\LARGE \bf 
    Bridging the Gap between Discrete Agent Strategies in Game Theory and Continuous Motion Planning for Dynamic Environments
    }

\usepackage{balance} 
\usepackage{amsmath,amssymb,amsfonts}

\usepackage{amsthm}
\usepackage{graphicx}
\usepackage{url}
\usepackage{multirow}
\usepackage{multicol}
\let\labelindent\relax
\usepackage{enumitem}
\usepackage{algorithm}
\usepackage{algpseudocode}
\usepackage[hidelinks]{hyperref}
\usepackage{color}
\usepackage{float}

\theoremstyle{definition}
\newtheorem{definition}{Definition}
\theoremstyle{remark}
\newtheorem{remark}{Remark}
\newtheorem{example}{Example}
\DeclareMathOperator{\argmax}{\text{argmax}}
\DeclareMathOperator{\argmin}{\text{argmin}}

\author{Hongrui Zheng$^{1}$, Zhijun Zhuang$^{1}$, Stephanie Wu$^{2}$, Shuo Yang$^{1}$, Rahul Mangharam$^{1}$
\thanks{$^{1}$ H. Zheng, Z. Zhuang, S. Yang, and R. Mangharam are with Department of Electrical and Systems Engineering, University of Pennsylvania, USA. $^{2}$ S. Wu is with the Department of Mathematics, University of Pennsylvania, USA. Correspond to {\tt\small hongruiz@seas.upenn.edu}.}
}

\begin{document}
\maketitle
\thispagestyle{plain}
\pagestyle{plain}

\begin{abstract}
    Generating competitive strategies and performing continuous motion planning simultaneously in an adversarial setting is a challenging problem. 
    In addition, understanding the intent of other agents is crucial to deploying autonomous systems in adversarial multi-agent environments. 
    Existing approaches either discretize agent action by grouping similar control inputs, sacrificing performance in motion planning, or plan in uninterpretable latent spaces, producing hard-to-understand agent behaviors.
    This paper proposes an agent strategy representation via \textbf{\textit{Policy Characteristic Space}} that maps the agent policies to a pre-specified low-dimensional space. 
    Policy Characteristic Space enables the discretization of agent policy switchings while preserving continuity in control. 
    Also, it provides intepretability of agent policies and clear intentions of policy switchings.
    Then, regret-based game-theoretic approaches can be applied in the Policy Characteristic Space to obtain high performance in adversarial environments.
    Our proposed method is assessed by conducting experiments in an autonomous racing scenario using scaled vehicles.
    Statistical evidence shows that our method significantly improves the win rate of ego agent and the method also generalizes well to unseen environments.
\end{abstract}
\section{Introduction}
Motion planning for autonomous agents in adversarial settings remains a challenging problem, especially for systems with continuous dynamics, where the system state, control action, and observation spaces are infinite. 
A natural solution is to adapt game-theoretic approaches for continuous motion planning~\cite{zanardi2021game}. The biggest challenge for this type of solution is to deal with continuity gracefully.

On the one hand, we could directly represent the game between agents without discretization. There exist models of games that incorporate continuity~\cite{sobel_continuous_1973}. Under these formulations with certain assumptions, the existence of equilibria can be proved; however, synthesizing practical strategies for agents is difficult, if not impossible.

On the other hand, we can discretize the output of motion planners and use them as actions in discrete games. 
Existing approaches deal with continuity in three ways: 1) discretizing actions by binning control input into intervals like bang-bang control, e.g., restricting the vehicle's turning command simply to three choices: turn left, turn right, and keep straight;
2) using a distribution-based policy to generate continuous control~\cite{schulman2017proximal, haarnoja2018soft}; 
3) formulating differential games by considering the game objective with the system differential equations~\cite{fridovich2020efficient, patil2023risk}.
In the first approach, bang-bang controllers lack control precision. They also produce abrupt changes between actions and often induce mechanical failure in the robotic system.
The second approach ends up generating similar controls from the first approach \cite{seyde_is_2021}. 
The third approach limits the game objective to being tightly coupled with the agent's system dynamics and is also computationally challenging in general as it involves solving constrained optimization problems.
Additionally, in all these three approaches, agent actions are not interpretable without external tools.

When understanding the opponent intent becomes crucial, existing approaches maintain a belief of the opponent, in the form of a distribution over some fixed parameterization of their policies, or a distribution directly over their actions given a game state~\cite{sinha_formulazero_2020,schwarting_stochastic_2021}. In practice, agent behaviors often deviate from expectation in such approaches, leading to suboptimal responses.


To address these challenges and provide practical insight into the use of game-theoretic approaches in continuous motion planning for agents in games, we propose modeling agent strategies in the \textbf{\textit{Policy Characteristics Space}}, which maps an instance of agent policies to a lower dimensional point in a pre-specified interpretable space (i.e., policy characteristic space).
We let agent \textit{policies} to produce actions that are continuous control inputs to the dynamic system, and let agent \textit{strategies} to select policies being used from a collection.
We also apply a regret minimization framework to show strategy optmization.

In this paper, we specifically study agent strategies for autonomous race cars in a head-to-head race. 
Autonomous racing provides a set of clearly specified metrics to balance safety and performance, making it a suitable scenario for research. Racing provides agents with the apparent goal of getting ahead of other agents and clear punishments if an agent crashes.
We address the following specific problem in this paper: 
\emph{Perform continuous motion planning in an adversary setting with approaches from game theory without sacrificing controllability while provides interpretability of agent actions.}



\subsection{Contributions}
This work has the following contributions:
\begin{enumerate}
    
    \item We encode agent strategies in the \textbf{Policy Characteristic Space} that allows for discrete actions in games while maintaining continuous control of the agent, and provide interpretable explanations of agent actions.
    We offline synthesize a population of policies lying in the Pareto Frontier of Policy Characteristic Space.
    \item We propose an online strategy optimization approach with approximated counterfactual regret minimization (CFR).
    \item We implement a pipeline that synthesizes agent policies offline with multi-objective optimization, and evaluate our proposed method in an autonomous racing scenario with realistic vehicle dynamics.
\end{enumerate}

\subsection{Related Work}
\subsubsection{Policies Population Learning}
There is a line of work in game theory that utilizes a collection of policies to create meta-policies, or meta-strategies. Our proposed approach is closest to Policy-Space Response Oracles (PSRO) and its variants~\cite{lanctot_unified_2017, balduzzi2019open, mcaleer2020pipeline, liu2022neupl}. PSRO also maintains ``meta-strategies" that select policies instead of actions. However, our approach differs fundamentally in the way in which the collection of policies is synthesized and organized. And our approach provides interpretability in agent decisions, whereas PSRO decisions are not explainable.

\subsubsection{Game in Latent Space}
Our proposed framework is also closely related to methods that operate in a latent space. Typically, encoder architectures are used to create implicit models of the world, agent intentions, agent interactions, or dynamics.
Hafner et al.~\cite{hafner_dream_2020} learns the latent dynamics of complex dynamic systems and trains agents in latent imagination for traditionally difficult control tasks.
Schwarting et al.~\cite{schwarting_deep_2021} uses latent imagination in self-play to produce interesting agent behaviors in racing games.
Xie et al.~\cite{xie_learning_2021} uses the latent space to represent the intention of the agent in multi-agent settings.
None of the above-mentioned approaches provides explainable latent spaces. In comparison, our proposed approach provides a foundation for interpretable spaces where agent actions are still abstracted into a lower dimension.

\subsubsection{Regret Minimization with Approximation}
Value function approximation~\cite{mnih_playing_2013} is widely used in reinforcement learning. Similarly, we approximate the converged counterfactual regret during the proposed approximated CFR.
Jin et al.~\cite{jin_regret_2018} approximates an advantage-like function as a proxy for regret in a single agent setting. Ours directly approximates the counterfactual regret in a multi-agent setting.
Brown et al.~\cite{brown_deep_2018} approximates the \emph{behavior or action} of CFR in a multi-agent setting without calculating and accumulating regrets. DeepCFR learns to approximate the best action on policy, while ours is trained off policy.




\section{Problem Setup}\label{sec:prob_setup}

We formally define our problem in Partially Observable Stochastic Games (POSG) \cite{hansen_dynamic_2004}. 
We consider a POSG given by the tuple $(\mathcal{I},\mathcal{S},\{b^0\}, \{\mathcal{A}_i\},\{\mathcal{O}_i\}, \{f_i\}, \{R_i\})$, where
\begin{itemize}
    \item $\mathcal{I}=\{1,\cdots,n\}$ is the finite set of agents;
    \item $\mathcal{S}=\times_{i\in\mathcal{I}}\mathcal{S}_i$ is the finite set of states and $\overrightarrow{s}=(s_1,\cdots,s_n)\in\mathcal{S}$ denotes a joint state;
    \item $b^0\in\Delta(\mathcal{S})$ represents the initial state distribution, and $\{b^0\}$ is the set of all possible initial state distributions;
    \item $\mathcal{A}_i$ is a finite set of actions available to agent $i$ and $\mathcal{A}=\times_{i\in\mathcal{I}}\mathcal{A}_i$ is the set of joint action (i.e., action profile) and $\overrightarrow{a}=(a_1,\cdots,a_n)\in \mathcal{A}$ denotes a joint action;
    \item $\mathcal{O}_i$ is a finite set of observations for agent $i$ and $\mathcal{O}=\times_{i\in\mathcal{I}}\mathcal{O}_i$ is the set of joint observations, where $\overrightarrow{o}=(o_1,\cdots,o_n)\in \mathcal{O}$ denotes a joint observation;
    \item $f_i: \mathcal{S}_i\times \mathcal{A}_i\rightarrow \mathcal{S}_i$ is the deterministic transition function (or system dynamics) of agent $i$;
    \item $R_i:\mathcal{S}\times\mathcal{A}\rightarrow\mathbb{R}$ is a reward function for agent $i$.
\end{itemize}

Next we define \textbf{Policies} and \textbf{Strategies} for agents.

\begin{definition}[Policy]\label{def:policy}
    We let $\pi_{i, \theta}(a_i | o_i)\in \Pi$ represent a \textbf{policy} parameterized by $\theta$ for agent $i$, and it is a distribution over all possible actions $a_i$ for the dynamical system of $i$ given the observation $o_i$.
\end{definition}
\begin{definition}[Strategy]\label{def:strat}
    We let $\rho_{i, \phi}(\pi_{i, \theta} | o_{i})$ represent a \textbf{strategy} (or meta-policy) parameterized by $\phi$ for agent $i$, and it is a distribution over possible policies $\pi_{i,\theta}$ given the observation $o_i$. 
    
\end{definition}

\begin{figure}[H]
    \centering
    \includegraphics[trim={0 1.5cm 0 2cm},width=0.98\columnwidth]{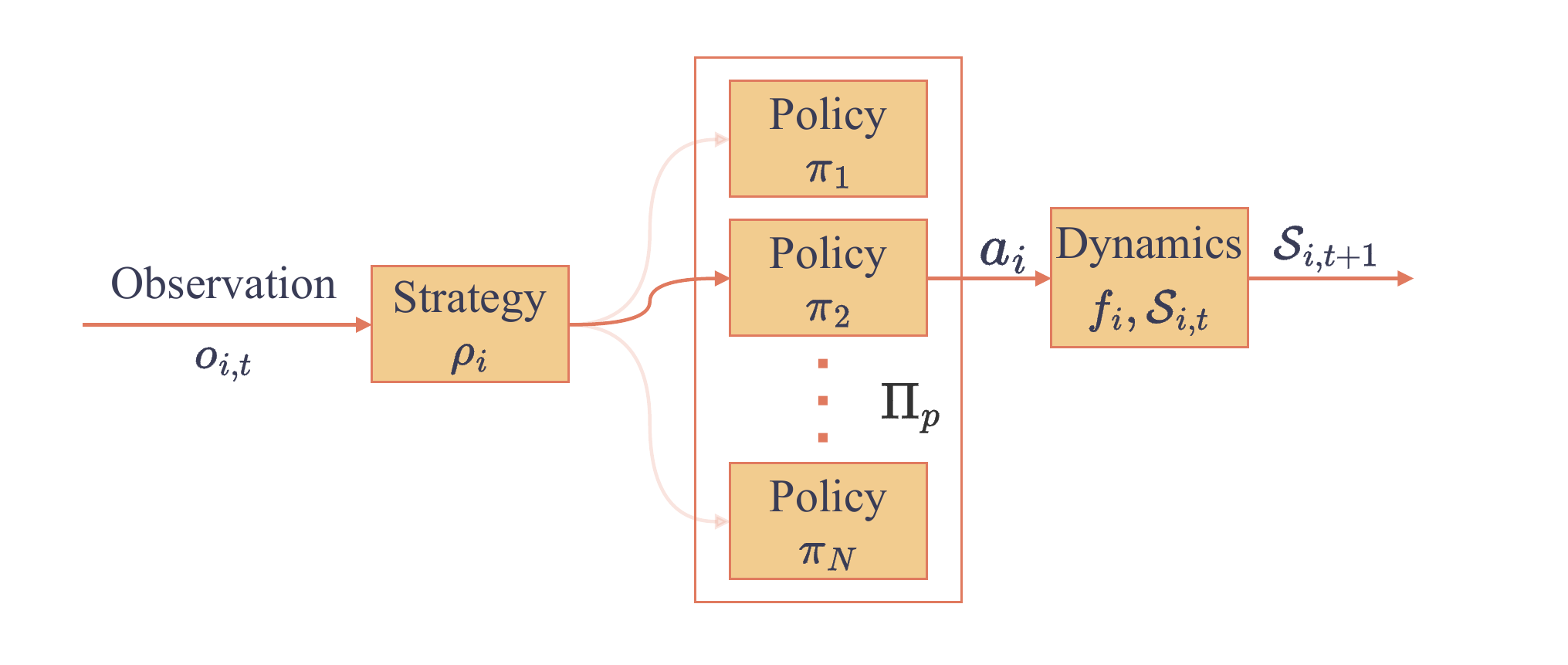}
    \caption{Hierarchical representation of strategies and policies. In our approach, strategies selects a policy $\pi$ out of a collection of policies $\Pi_p$. Then policies generate actions $a_i$ (control inputs) for the agent's dynamic system.
    }
    \label{fig:strat_pol}
    \vspace{-10pt}
\end{figure}

We illustrate the hierarchical relationship among strategies, policies, and the agent's system dynamics in Figure \ref{fig:strat_pol}.
In this work, we are interested in developing some strategies for autonomous agents with given tasks.
To better understand the strategy development, we define the policy characteristic space now.


\begin{definition}[Policy Characteristic Space]\label{def:pcs}
    A policy characteristic function $g_c(\pi_i):|\Pi|\rightarrow \mathbb{R}$ describes a characteristic $c_{\pi_{i}}$ of the policy $\pi_i$.
    In addition, we define a set of these functions as $g_c \in \mathcal{G}$, where each $g_c$ calculates a different characteristic.
    A vector of characteristic values of the policy $\pi_i$ is denoted as $\bar{c}_{\pi_i}\in\mathbb{R}^{|\mathcal{G}|}$, where each element of $\bar{c}_{\pi_i}$ corresponds to a characteristic function in $\mathcal{G}$.
    Using all functions in $\mathcal{G}$ as basis functions, the space spanned by multiple policy characteristic functions is referred to as the \textbf{\textit{Policy Characteristic Space}} (PCS). $\bar{c}_{\pi_i}$ serves as the coordinates for a policy $\pi_i$ in the PCS.
\end{definition}

\begin{example}\label{rem:ex_char}
One can define policy characteristics functions $g_c$ as functions that output a real number.
For instance, in autonomous racing scenario, we can define two functions $g_{c, 1}$ and $g_{c, 2}$ where $g_{c, 1}(\pi)$ measures the safety level of policy $\pi$ and $g_{c, 2}(\pi)$ measures the performance level of $\pi$.
\end{example}

\begin{remark}\label{rem:pcs}
It is hard to compute the true characteristics value of a given policy under all possible scenarios it might face.
However, we can use a sampling-based method to estimate its value.
We denote a state-space trajectory of agent $i$ produced by a policy $\pi_i$ as $\mathcal{T}_i=\{s_{i,t}\}_{t=[1,\cdots,T]}$ and a set of $N$ such trajectories generated by $\pi_i$ as $\left\{\mathcal{T}_i^1,\ldots,\mathcal{T}_i^N\right\}$. 
Then, to measure a characteristic $g_c(\pi_i)$ in PCS, we estimate it through the trajectories set induced by $\pi_i$, i.e., we let $c_{\pi_i}=g_c(\pi_i)=g_c\left(\left\{\mathcal{T}_i^1,\ldots,\mathcal{T}_i^N\right\}\right)$ with a slight abuse of notations. Finally, to map a policy to the PCS, we calculated all policy characteristics values $\bar{c}_{\pi_i}$ for $\pi_i$.
\end{remark}

During \textbf{offline phase}, our objective is to synthesize a population of policies with diverse characteristic values.
To do that, we aim to contain all policies on the \emph{Pareto Frontier}.
We formally define our offline objective as the following:
\begin{equation}\label{eq:synth_objective}
\begin{split}
    \text{Find}~~~ & \Pi_p\subset \Pi \\
    \operatorname{\text{subject to}~~~} & \forall \pi_j\in \Pi\backslash \Pi_p, \exists \pi_i\in \Pi_p, \text{s.t. } \bar{c}_{\pi_j} \prec \bar{c}_{\pi_i}
\end{split}
\end{equation}
where $\bar{c}_{\pi_j} \prec \bar{c}_{\pi_i}$ denotes that policies in $\Pi_p$ strictly dominates (always preferred over) all other policies, thus $\Pi_p$ is the pareto-optimal policy set.

\begin{example}(Cont.)
    Again, consider the autonomous racing example with safety ($g_{c, 1}$) and performance ($g_{c, 2}$) characteristics. 
    One can say $\bar{c}_{\pi_i} \prec \bar{c}_{\pi_j}$ if the policy $\pi_i$ has characteristic value $\bar{c}_{\pi_i}=[0.5\;0.5]$ and $\pi_j$ has $\bar{c}_{\pi_j}=[0.7\;0.6]$; however, it is not this case if $\bar{c}_{\pi_j}=[0.7\;0.4]$.
\end{example}

In the \textbf{online phase}, we develop the strategy whose support is the policy Pareto Frontier $\Pi_p$ computed from~\eqref{eq:synth_objective} offline.
During online phase, the objective is to find the strategy that maximizes the accumulated reward for the agent $i$ that we control. 
We denote the policy selected by strategy $\rho_{i,\phi}$ at each step $\tau$ for agent $i$ as $\pi_{i,\tau}^\phi$. 
For example, under a strategy parameterized by $\phi$, the policy selected by a \textit{greedy} selection rule can be as follows:
\begin{equation} \label{eq:policy_select}
    \pi_{i,\tau, g}^{\phi} = \underset{\pi_{i, \theta}}{\operatorname{argmax}}\, \rho_{i,\phi}(\pi_{i, \theta} | o_{i,\tau}).
\end{equation}
Suppose ego agent is $i$ and all other agents have policy profile $\pi_{-i}$. 
Over a horizon $T$, the optimal strategy for agent $i$ is then formulated as follows:
\begin{equation} \label{eq:objective}
    \rho_i^* = \underset{\rho_{i,\phi}}{\operatorname{argmax}}\underset{t=1}{\overset{T}{\sum}}R_{i}(\overrightarrow{s_t}, \pi^{\phi}_{t}(\overrightarrow{o_{t}})), \text{where } \pi^{\phi}_{t}=(\pi^{\phi}_{i, t}, \pi_{-i}).
\end{equation}
In other words, the online objective for strategy optimization is to find the strategy that maximizes the sum of returns over the horizon $T$, and at each step of the horizon, the strategy selects a policy for the agent.

\begin{remark}
    Note that we consider finite-horizon undiscounted return as the objective in this work, but one can also consider infinite-horizon discounted return and our framework is still applicable.
\end{remark}

To account for interpretability, we use the greedy selection rule in~\eqref{eq:policy_select} so the used policy at each time step is a ``pure policy".
Thus, our online optimization problem is to find the \emph{optimal policy swtiching sequence} where policies come from the offline synthesized Pareto-optimal policy set in~\eqref{eq:synth_objective}.
We describe our proposed method in the coming section.
\section{Methodology}

\begin{figure*}[]
    \centering
    \includegraphics[trim={0 1cm 0 1cm},width=1.8\columnwidth]{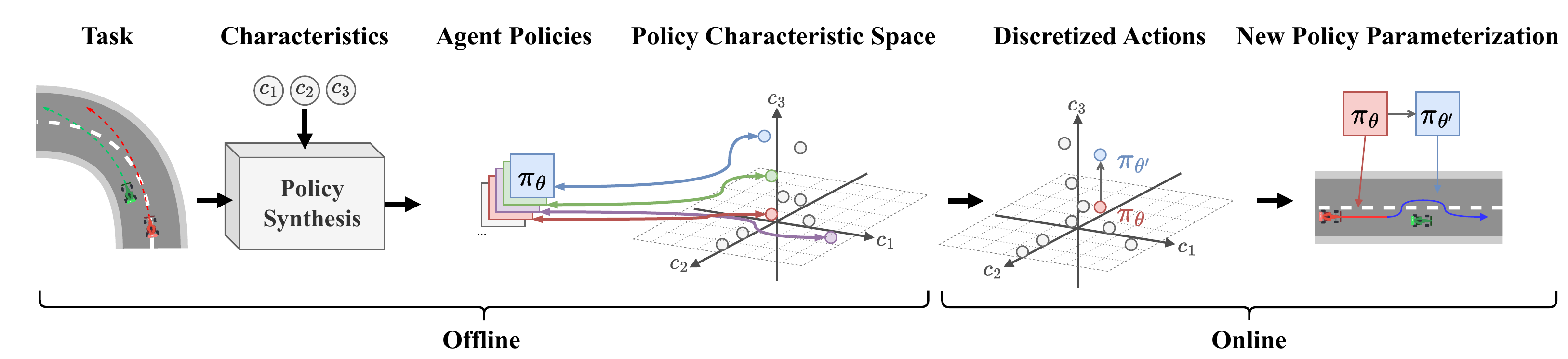}
    \caption{
    Overview: During offline phase, agent policies are first synthesized based on the given task and policy characteristics functions. Agent strategies are modeled as a collection of policies in the policy characteristics space. Then, during online phase, strategies perform discretized actions by switching between discovered policies based on their characteristics values.}
    \label{fig:overview}
    \vspace{-10pt}
\end{figure*}

In the following section, we describe our approach to solving the offline policy synthesis problem in Equation~\eqref{eq:synth_objective}, and the online strategy optimization problem in Equation~\eqref{eq:objective} in Section \ref{sec:prob_setup}.
An overview of the proposed method is shown in Figure \ref{fig:overview}. 
Given a task and desired characteristics of policies, we first synthesize a population of compatible agent policies, represented in the PCS as points. 
Then we optimize the online strategy that controls the selection of policies to maximize the total returns. With our proposed method, we allow for discretization at the strategy level without losing continuous control at the policy level.

\subsection{Policy Synthesis via Multi-objective Optimization}\label{sec:offline}
\begin{figure}[h]
    \centering
    \includegraphics[trim={2cm 1cm 0 0},width=0.9\columnwidth]{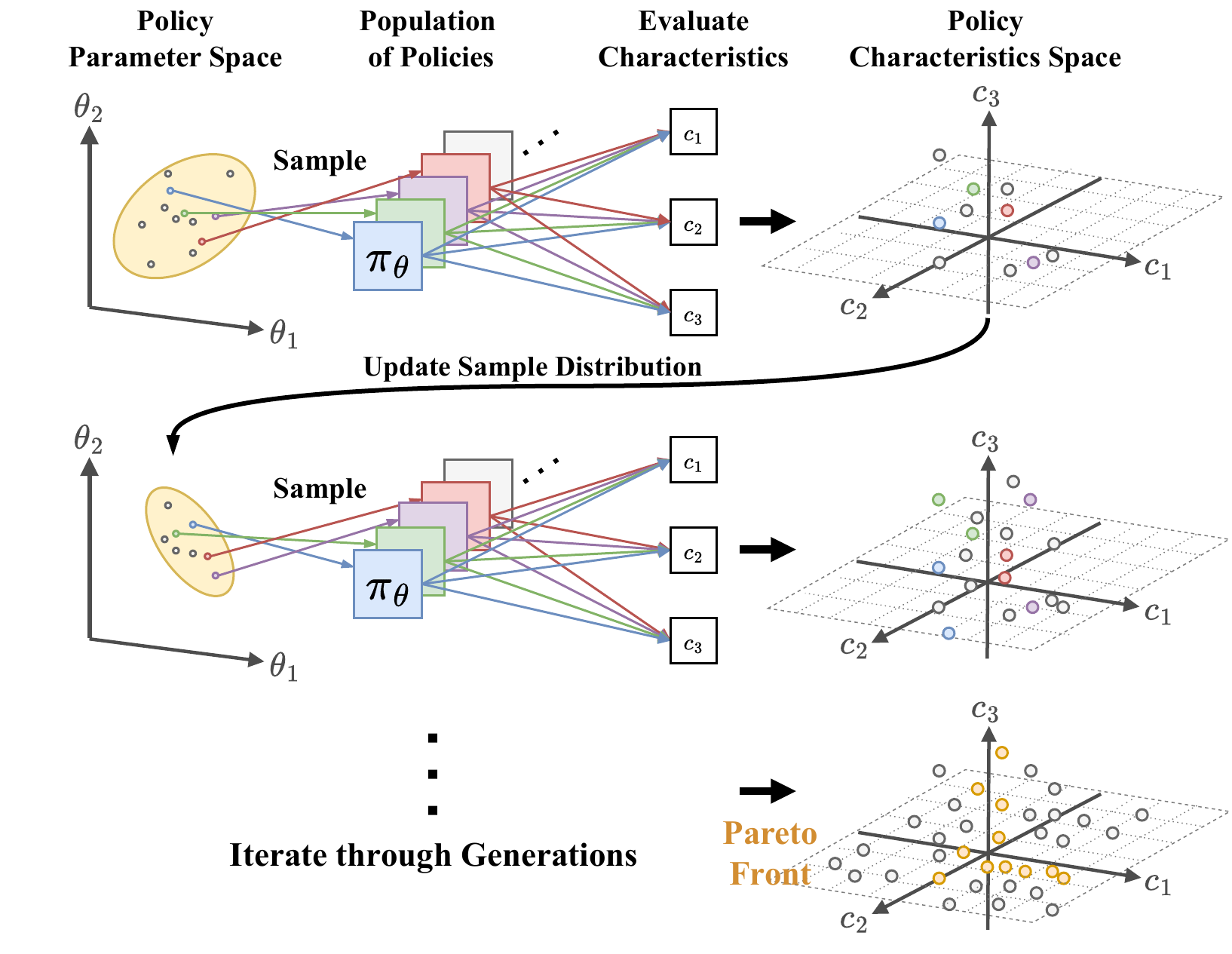}
    \caption{Population Synthesis: Population-based optimization is an iterative process that maintains a sampling distribution. At each iteration, policies parameterizations are sampled using the distribution, then evaluated by policy characteristics functions. Based on the characteristics value of each policy, the sampling distribution is updated.}
    \label{fig:pop_synth}
\end{figure}

To create $\Pi_p$ in Equation~\eqref{eq:synth_objective}, we use a population-based algorithm to find the policies on the Pareto Frontier across different axis of the Policy Characteristics Space.
The population synthesis process is depicted in Figure \ref{fig:pop_synth}. Specifically, we use the Multi-Objective Covariance Matrix Adaptation Evolution Strategy (MO-CMA-ES) \cite{hansen_cma_2016} to create new policies. We describe the algorithm in Algorithm \ref{alg:synth}.

\begin{algorithm}
    \caption{MO-CMA-ES}\label{alg:synth}
    \begin{algorithmic}[1]
        \Function{MO-CMA-ES}{$\texttt{eval}, n, \textit{iter}$} 
        \State $\textbf{Initialize}~~~ \lambda, \mu, \sigma, C=I, p_\sigma=\bar{0}, p_c=\bar{0}, \chi=\emptyset$ 
        \For{$\textit{gen} \gets 1~\textbf{to}~\textit{iter}$}  
            \For{$i \gets 1~\textbf{to}~n$} 
                \State $\theta_i \gets \texttt{sample}(\mu,\sigma^2C)$ 
                \State $\bar{c}_i \gets \texttt{eval}(\pi_{\theta_i})$ 
            \EndFor
            \State $l_{1,\ldots,n} \gets \texttt{hypervolume\_loss}(\chi, \bar{c}_{1,\ldots, n})$
            \State $\theta_{s(1),\ldots,s(n)} \gets \texttt{sort}(\theta_{1,\ldots,n}, l_{1,\ldots,n})$
            \State $\chi \gets \texttt{add}(\chi, \theta_{s(1), \ldots, s(\lambda)})$
            
            \State $p_\sigma, p_c \gets \texttt{update\_stepsize}(p_\sigma, p_c, \chi, \sigma, C)$ 
            \State $C, \sigma \gets \texttt{update\_covariance}(p_\sigma, p_c, \sigma, C)$ 
            \State $\mu \gets \texttt{update\_mean}(\chi)$ 
            
        \EndFor
        \State $\textbf{return}~ \chi\textit{.pareto}$ 
        \EndFunction
    \end{algorithmic}
\end{algorithm}


In line 2, we initialize the elite ratio $\lambda$, the multivariate normal distribution we sample from, set the isotropic and anisotropic evolution paths to zero vectors, and initialize the population archive as an empty set.
Line 3 initiates the generation loop.
In lines 4 to 7, we sample $n$ policy parameters $\theta_i$ from the current sampling distribution, and evaluate each of them to retrieve the policy characteristics value vectors $\bar{c}_i$.
In line 8, we evaluate the hypervolume loss based on a unary hypervolume indicator \cite{fonseca_improved_2006} for each individual sampled. This loss indicates how much an individual expands the hypervolume containing the population $\chi$ in the optimization directions.
In line 9, we sort the sampled individuals based on their loss values.
In line 10, we add the top $\lambda$ sampled individuals to the population archive $\chi$.
In line 11, we update the evolution paths that control how much we should shrink or inflate the covariance based on the updated population.
In lines 12 and 13, we update the sampling distribution based on the updated evolution paths and population archive.
Finally, we return the Pareto Frontier of the population archive as our collection of policies.

Now, we describe how strategies specifically enable switching between policies in the PCS.
We define strategy outputs as $\epsilon\cdot e_k$, where $\epsilon\in \mathbb{R}$ and $e_k$ is a vector with length $|\mathcal{G}|$ where the $k$-th element is 1 and all other elements are 0.
For instance, for any point $\bar{c}_i$ (i.e., a policy) in the PCS, the new point $\bar{c}_j$ (i.e., a new policy) induced by $\epsilon e_{k+}$ is $\bar{c}_j=\bar{c}_i+\epsilon e_k$, and $\epsilon e_{k-}$ is $\bar{c}_j=\bar{c}_i-\epsilon e_k$.
In other words, an action by the strategy changes the value of only one of the policy characteristic values.
However, it might happen that the switched policy $\bar{c}_j$ does not correspond to any policy in the Pareto-optimal policy set $\Pi_p$,
we choose the nearest point of  $\bar{c}_j$ in the $\Pi_p$ in that case, i.e., the new policy is actually selected by
\begin{equation}\label{eq:act_2_pol}
    \pi_{\theta_{\text{new}}}=\underset{\pi_{\theta}\in \Pi_p}{\argmin}||(\bar{c}_i + \epsilon e_k)-\bar{c}_{\pi_\theta}||_2.
\end{equation}
In our case study, we chose $\epsilon$ to be a fixed value for simplicity. In practice, multiple values could be used and doing so will increase the number of possible strategy outputs.
Thus, such a transformation reduces the motion planning action space from infinitely large to strategy action space of $2{|\mathcal{G}|}$, and also provides an explanation for agent actions, depending on how these characteristic functions are defined.


As explained in Remark \ref{rem:pcs}, in order to provide characteristics general across different scenarios an agent might encounter, the \texttt{eval} function in Algorithm \ref{alg:synth} is designed to collect results from many rollouts. In Section \ref{sec:exp_offline}, we detail the evaluation function used in our case study.

\subsection{Strategy Optimization via Counterfactual Regret Minimization}\label{sec:online}
In the following section, we describe the formal game theory model used for strategy optimization, and the Counterfactual Regret Minimization algorithm using this formal model.
\subsubsection{Game Model} 
In the game we designed, two autonomous race cars start side by side and race for two laps. The winner is awarded by the lead it had at the end of the race, and the loser is penalized by the negative of the same value. Any collision results in no reward for either agent. Formally, we define the game as a \textbf{two-player zero sum extensive form game} with partial information and perfect recall.
With our definition of strategy, we will be able to establish a discrete game model without having to explicitly describe the continuous dynamics.
All actions taken by the agents (strategy) in the game model are switching policies; therefore, the timesteps are equivalent to the timesteps defined in Definition \ref{def:strat}.
We denote all agents except $i$ as $-i$.
It is important to note that the policy characteristic value vectors $\bar{c}_{\pi_{-i}}$ of other agents can be calculated using only their state space trajectories. We include them in observations at every game step.
We define the history $h_{i,\tau}$ at time $\tau$ for agent $i$ as $h_{i,\tau}=\left\{\left(\bar{c}_{\pi_{-i},1}, \ldots, \bar{c}_{\pi_{-i}, \tau-1}\right), \left(\bar{c}_{\pi_i, 1}, \ldots, \bar{c}_{\pi_i, \tau-1}\right)\right\}$.
A history vector of all agents at $\tau$ is $h_{\tau}$, and the history vector excluding agent $i$ is $h_{-i,\tau}$.
Our game terminates when $\tau=T$, and a terminal history of agent $i$ is denoted as $z_i\in Z_i$, where $Z_i$ is all possible terminal histories.
The terminal utility, or payoff, for agent $i$ at a terminal history $z_i$ is denoted as $\mu_i(z_i)$.
Following Definition \ref{def:strat}, the strategy of the agent $i$ is denoted as $\rho_i$, and $\rho_{-i}$ is strategies of all agents except $i$.
We denote $\rho$ the strategy profile that comprises all the strategies of the agents. 
The probability of reaching a specific history vector $h$ with strategy profile $\rho$ is denoted as $Pr^{\rho}(h)$.
And $Pr^{\rho}_{-i}(h)$ denotes the probability of reaching $h$ without the contribution of the agent $i$. 

\subsubsection{Admissible Actions Set}
In our case study, we let $\epsilon$ in Equation~\eqref{eq:act_2_pol} to be a fix value, hence the set of admissable actions for an agent is then the discrete set $\mathcal{E}=\{e_{1+}, e_{1-}, e_{2+}, e_{2-}, \ldots, e_{|\mathcal{G}|+}, e_{|\mathcal{G}|-}\}$. We will denote an action in the game model by $e_k\in\mathcal{E}$.

\subsubsection{Counterfactual Regret Minimization}
The counterfactual value $\mathbf{v}_i^\rho(h)$ of agent $i$ is the expected utility it receives when reaching $h$ with probability one:
\begin{equation}
    \label{eq:cf_val}
    \mathbf{v}_{i}^\rho(h)=\sum_{z \in Z_{h}} Pr^{\rho}_{-i}(z[h]) Pr^{\rho}(z[h]\rightarrow z) \mu_{i}(z),
\end{equation}
where $Z_h$ is the set of all terminal histories reachable from $h$ and $z[h]$ is the prefix of $z$ up to $h$ and $Pr^{\rho}_{-i}(z[h])$ is the probability of reaching $z[h]$ without agent $i$'s contribution, where ``without contribution" means that the whole game state will reach $z[h]$ regardless of agent $i$'s actions.
$Pr^{\rho}(z[h]\rightarrow z)$ is the probability of reaching $z$ from $z[h]$. 
The immediate or instantaneous counterfactual regret for action $e_k$ at step $\tau$ is
\begin{equation}
    \label{eq:imm_cf}
    r_i^\tau(h, e_k) = \mathbf{v}_{i}^{\rho^\tau}(h, e_k) - \mathbf{v}_{i}^{\rho^\tau}(h),
\end{equation}
where $\mathbf{v}_{i}^\rho(h, e_k)$ follows the same calculation as counterfactual value and assumes that agent $i$ takes action $e_k$ at the history $h$ with probability one.
The \textit{counterfactual regret} for history $h$ and action $e_k$ at step $\tau$ is
\begin{equation}
    \label{eq:cfr}
    \mathcal{R}_i^\tau(h, e_k)=\sum_{t=1}^\tau r^{t}(h, e_k).
\end{equation}
Intuitively, the counterfactual regret measures how much more utility each action taken by the agent $i$ results in at the end of the game over the expected utility. 
Hence, choosing the action that has the highest counterfactual regret will result in more utility received by the agent at the end of the game.
Next, we introduce the function approximator $g$ for the counterfactual regret at iteration $\tau$ as:
\begin{equation}\label{eq:approx_cfr}
    \mathcal{R}_i^\tau(h, e_k) \approx \mathcal{R}_{\operatorname{approx}}(h, e_k).
\end{equation}
Additionally, we clip the counterfactual regret by using $\mathcal{R}_{i,+}^\tau(h, e_k)=\max\{\mathcal{R}_i^\tau(h, e_k), 0\}$. 
The clipping can be achieved by using ReLU activation in $\mathcal{R}_{\operatorname{approx}}$.
Finally, we choose the action with the highest approximate counterfactual regret with probability one:
\begin{equation}
\begin{split}
    \label{eq:cf_strat}
    e_{k,\tau+1} &= \underset{e_{k,\tau+1}}{\argmax}~ \mathcal{R}_{i,+}^{\tau+1}(h, e_{k,\tau+1}) \\
    &\approx\underset{e_{k,\tau+1}}{\argmax}~ \mathcal{R}_{\operatorname{approx}}(h, e_{k,\tau+1}).
\end{split}
\end{equation}
If the sum of the counterfactual regret of all actions at a step is zero, then any arbitrary action may be chosen at that step.
The above action $e_{k,\tau+1}$ is the action defined in Policy Characteris Space and it guides the current policy to switch the next policy.
\begin{remark}
    Note that one can also choose Regret Matching to select actions~\cite{zinkevich_regret_2007}, i.e., the selection probability for each action is proportional to its counterfactual regret value.
    We directly choose the action with the highest regret in this work due to its better practical performance~\cite{brown_deep_2018}.
\end{remark}
\section{Experiments}
In our experiments, we aim to answer the following three questions.
\begin{enumerate}
    \item Does being game-theoretic improve an agent's win rate against a competitive opponent?
    \item Does the proposed agent action discretization provide interpretable explanations for agent actions?
    \item Does the proposed approach generalize to unseen environments and unseen opponents?
\end{enumerate}
In the following sections, we will first introduce how we set up the autonomous racing task, how we parameterize policies with a motion planner, how we implement strategy optimization, and our experimental results.

\subsection{Simulation Setup}\label{sec:sim_setup}
We study a two-player head-to-head autonomous race scenario as a case study. The simulation environment~\cite{okelly_f1tenth_2020} is a gym~\cite{brockman_openai_2016} environment with a dynamic bicycle model~\cite{althoff_commonroad_2017} that considers side slip. The objective of the ego in this game is to progress further along the track than the opponent in the given amount of time without crashing into the environment or the other agent. The state space of an agent in the environment is $x_i = [x, y, \psi, s]$ where $x, y$ is the position of the agent in the world, $\psi$ is the heading angle, and $s$ is the longitudinal displacement along the track in the Frenet coordinate system. The control input space of an agent is $u_i=[\delta, v]$ where $\delta$ is the steering angle and $v$ is the desired longitudinal velocity.
Additionally, each agent receives a range measurement vector produced by a ray-marching LiDAR simulation $\mathbf{r}\in\mathbb{R}^q$, where $q$ is the number of laser beams, and the pose of other agents nearby.
We design the reward function based on the lead along the track that the winner of the race has. Consequently, the utility $\mu$ in the game model is $\mu_{w} = (s_{w} - s_{l})$, and $\mu_{l} = -\mu_{w}$ to make the game zero sum. $s_w$ and $s_l$ are the longitudinal displacement coordinate of the winner and the loser at the end of the race, respectively.
If there are any collisions that end the game prematurely, both agents get zero utility.
Each game between two agents runs for a fixed number of game steps $\tau=[1,\ldots,T]$, and each game step $\tau$ consists of a fixed number of simulation timesteps $t=[1,\ldots,\Gamma]$.

\subsection{Offline Phase: Motion Planning and Policy Synthesis}\label{sec:exp_offline}
\begin{figure}[b]
    \centering
    \includegraphics[trim={0cm 0cm 0cm 2cm},width=0.9\columnwidth]{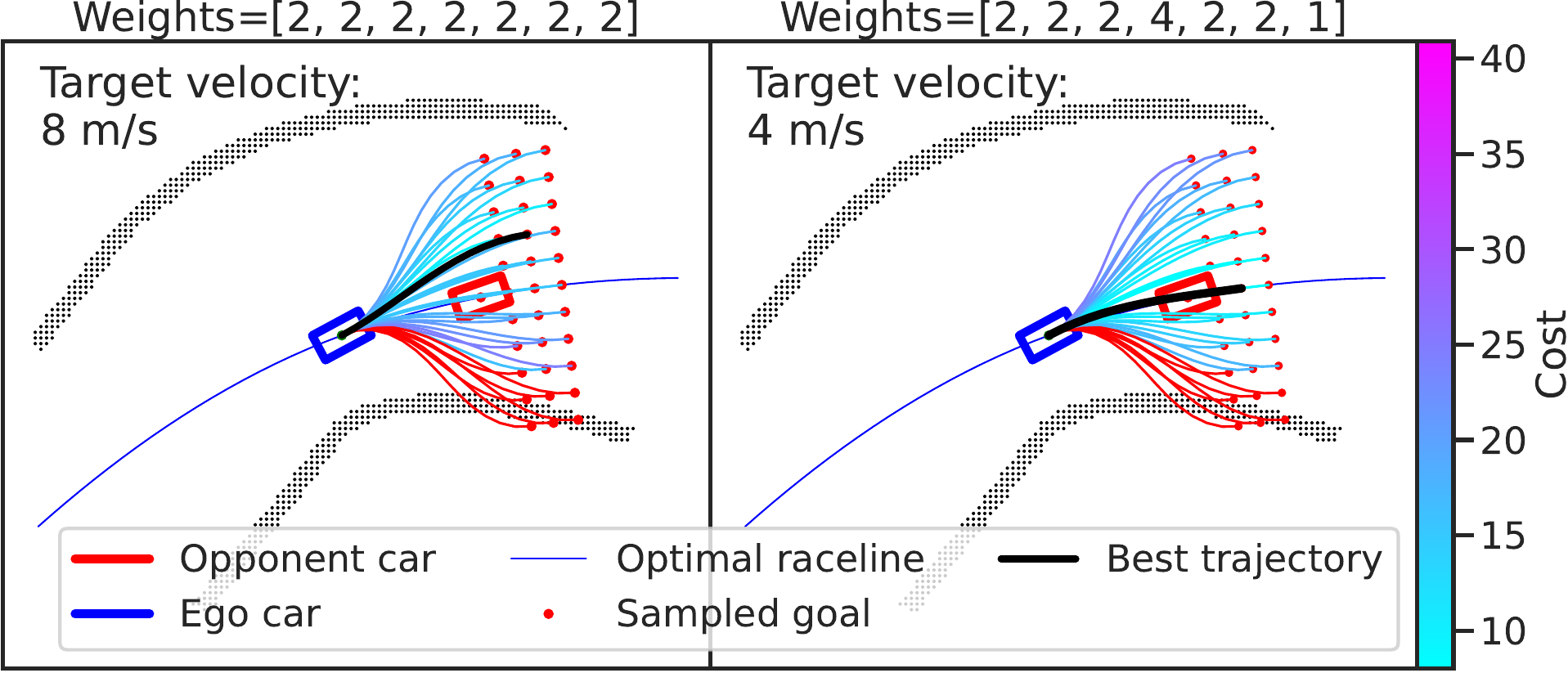}
    \caption{Effect of the different weighting of cost functions on agent behavior. The red trajectories are in collision with the track, thus assigned infinite cost.}
    \label{fig:planner}
\end{figure}
In Definition \ref{def:policy}, we defined that a policy takes in observations from the environment and produces control inputs for the dynamical system of agents. 
In our case study, agents use a motion planner as a policy. 
In general, the policy could be anything that can be consistently parameterized by a vector (e.g. weights of a neural network) and produces control inputs.

We use a sampling-based motion planner. First, the planner samples local goals for the vehicle in a lattice pattern, as shown in Figure \ref{fig:planner}. Then, dynamically feasible trajectories are generated~\cite{kelly_reactive_2003} that take the vehicle from the current state to the sampled goal state.
For each sampled candidate trajectory, we calculate a few cost functions that correspond to geometric properties (e.g. maximum curvature, deviation from the race line, etc.). The trajectory for the agent to follow is the candidate trajectory that has the lowest weighted sum of all cost function values. We parameterize the motion planner with the weight vector. More details on the cost functions used can be found in Appendix \ref{app:costs}. After selecting a trajectory, the vehicle control input is generated using Pure Pursuit~\cite{noauthor_implementation_nodate}.
We can see the different behavior of the agent induced by having different cost weights in Figure \ref{fig:planner}. The planner on the left weighs the penalty on deviating from a predetermined raceline less, thus generating a path that keeps a higher velocity around the opponent vehicle. The planner on the right weights this penalty more, thus generating a path that slows down and follows the opponent in front, while keeping a trajectory closely matching the raceline.

For the racing task, we define two policy characteristic functions, aggressiveness and restraint of the agents. Recall from Remark $\ref{rem:pcs}$, these functions take in set of agent state space trajectories, we define the following policy characteristic functions:
\vspace{-5pt}
\begin{equation}
\begin{split}
    g_{\text{agg}}(\{\mathcal{T}_i^1,\ldots,\mathcal{T}_i^N\}) &= \frac{1}{N}\sum_{j=1}^{N} (s_{j,\text{ego}} - s_{j, \text{opp}})\\
    g_{\text{res}}(\{\mathcal{T}_i^1,\ldots,\mathcal{T}_i^N\}) &= \frac{1}{N}\sum_{j=1}^{N} \left(-\frac{1}{T} \sum_{t=1}^{T} \min_q \left[\frac{\mathbf{r}_{j,t,q}}{\dot{\mathbf{r}}_{j,t,q}}\right]_{+\infty}\right)
\end{split}
\end{equation}
$g_{\text{agg}}$ evaluates a policy's aggressiveness with the ego's average progress along the track over the opponent. $g_{\text{res}}$ evaluates a policy's restraint with the average instantaneous time to collision (iTTC) calculated using the laser scan measurements.
$N$ is the number of evaluated trajectories, $T$ is the horizon length of the trajectories, $\dot{\mathbf{r}}$ is the time derivative of range measurement, and $[]_{+\infty}$ is the operator that sets negative elements to infinity.
We set up the \texttt{eval} function in MO-CMA-ES (Algorithm \ref{alg:synth}) by first selecting a fixed number of random sections of the race track and starting positions and a set of random policies as opponents. In each call to \texttt{eval}, we pit each sampled policy against all pairings of environment and opponent to collect the set of trajectories. Finally, we use $g_{\text{agg}}$ and $g_{\text{res}}$ as the returns of $\texttt{eval}$ to obtain the collection of policies $\Pi_p$.

\subsection{Online: Strategy Opitmization with Approximated CFR}\label{sec:exp_cfr}

Before online strategy optimization, we need to first train the regret approximator $\mathcal{R}_{\operatorname{approx}}(h, e_k)$ that estimates the counterfactual regret for each action given a history at the final iteration of CFR.
Training data are collected by playing full extensive games represented as game trees.
First, we allow each agent to take $m$ actions in total in a game.
The initial starting policies for both agents in the root node are chosen as random policies from $\Pi_p$.
Then, we traverse every single branch on the game tree by taking all combinations of action at each node for both agents. At the terminal nodes of the tree, the final utilities are calculated for each game, and the corresponding counterfactual regret is also calculated for every action for every corresponding history present on the tree.
The total number of games played between two agents is $N_{\text{init}}^2(2{|\mathcal{G}|})^{2m}$, where $N_{\text{init}}$ is the number of initial starting policies for each agent. For more details on how subsets of policies used are selected, and the size of the training dataset, please refer to Appendix \ref{app:nn}. The number of games played, even considering the relatively small number of $N_{\text{init}}$ and $m$, is still massive. This further strengthens the need for an approximator for counterfactual regret.

Online, the game-theoretic planner works in the following order. First, the ego selects a random starting policy from $\Pi_p$. Then, it observes the opponent's trajectory, calculates the corresponding policy characteristic values for the opponent, and predicts the counterfactual regret for each available action. Next, the action with highest approximate counterfactual regret is taken, and moves the ego's current position in the PCS to a new point by switching policies. The corresponding cost weights for the motion planner are used to update the motion planner. Lastly, the motion planner generates the control input for the ego agent.

\subsection{Exeprimental Results}
\subsubsection{Being game-theoretic improves win rate against other agents}
\begin{table}[h]
\vspace{-5pt}
\centering
\caption{Win rates in head-to-head racing experiments with mean win rate differences and p-values.}
\begin{tabular}{|c|c|c|c|c|}
\hline
\multirow{3}{*}{Opponent} & \multicolumn{2}{c|}{Ego} & \multirow{3}{*}{$\Delta_\mu$} & \multirow{3}{*}{p-value} \\
\cline{2-3}
 & Win rate & Win rate & & \\
     & Non-GT & GT & & \\
\hline 
\multicolumn{5}{|c|}{On Seen Map} \\
\hline
Non-GT & $0.515 \pm 0.251$ & $0.569 \pm 0.213$ & $0.054$ & 0.0142 \\
Random & $0.624 \pm 0.225$ & $0.670 \pm 0.199$ & $0.046$ & 0.00370 \\
Unseen & $0.586 \pm 0.101$ & $0.597 \pm 0.089$ & $0.011$ & 0.0863 \\
\hline
\multicolumn{5}{|c|}{On Unseen Map} \\
\hline
Non-GT & $0.553 \pm 0.256$ & $0.628 \pm 0.180$ & $0.075$ & 0.0124 \\
Random & $0.625 \pm 0.278$ & $0.738 \pm 0.172$ & $0.113$ & 0.00276 \\
Unseen & $0.556 \pm 0.101$ & $0.565 \pm 0.098$ & $0.009$ & 0.147 \\
\hline
\end{tabular}
\label{tab:online_exp}
\end{table}

To answer the first question, and in order to show the effectiveness of game-theoretic planning, we race the policies from our framework against various opponents in different environments. 
Each experiment is a single head-to-head race with a fixed number of game steps. We designate the winner of the race as the agent who progressed further down the track at the end of the duration. 
To ensure fairness, the agents start side by side at the same starting line on track and alternate starting positions. The starting line is also randomized five times for one pair of agents. 
There are three types of agents in the experiment. 
The \textbf{GT agent} starts with a random policy in $\Pi_p$ and uses game-theoretic planning online, switching between different policies in $\Pi_p$. 
The \textbf{non-GT agent} also starts with a random policy in $\Pi_p$ but does not change its policy online. 
The \textbf{random agent} is a random selection from all the policies explored during policy synthesis (not necessarily in $\Pi_p$), and does not change its polciy online. 
Lastly, the \textbf{unseen agent} is a winning motion planner from an annual autonomous racing competition that allows parameterization \cite{f1tenth_f1tenth_nodate, zhijunzhuang_zzjun725f1tenth-racing-stack-icra22_2024}. 
The ego agents in the experiments are GT and non-GT agents, while the opponents are non-GT, random, and unseen (Table \ref{tab:online_exp}).

We choose 20 different variants of each agent. 
Thus, each table cell in Table \ref{tab:online_exp} consists of statistics from $20^2 \times2\times 5=4000$ head-to-head racing games. 
Each agent is allowed $m=4$ actions in the game, with each step lasting $8$ seconds. 
The total length of the games is $40$ seconds, with the first $8$ seconds as the initialization to observe the opponent.
We report the results of paired t-tests against all opponents with the null hypothesis that the use of the proposed game-theoretic planner does not change the win rate.
As shown in Table \ref{tab:online_exp}, the main result of this experiment is that \emph{the p-value is small enough to reject the null hypothesis in most cases}. Across all pairings of ego and opponent, there is an increase in average win rate by using our proposed approximated CFR, significantly in most cases. Although not as significant when playing an unseen opponent, the trend is still present. This finding validates the effect of our game-theoretic planner by showing a significant improvement in the win rate.

\subsubsection{Using policy characteristic space provides interpretable agent actions}
\begin{figure}[h]
    \centering
    \includegraphics[trim={2cm 1cm 1cm 0cm},width=0.98\columnwidth]{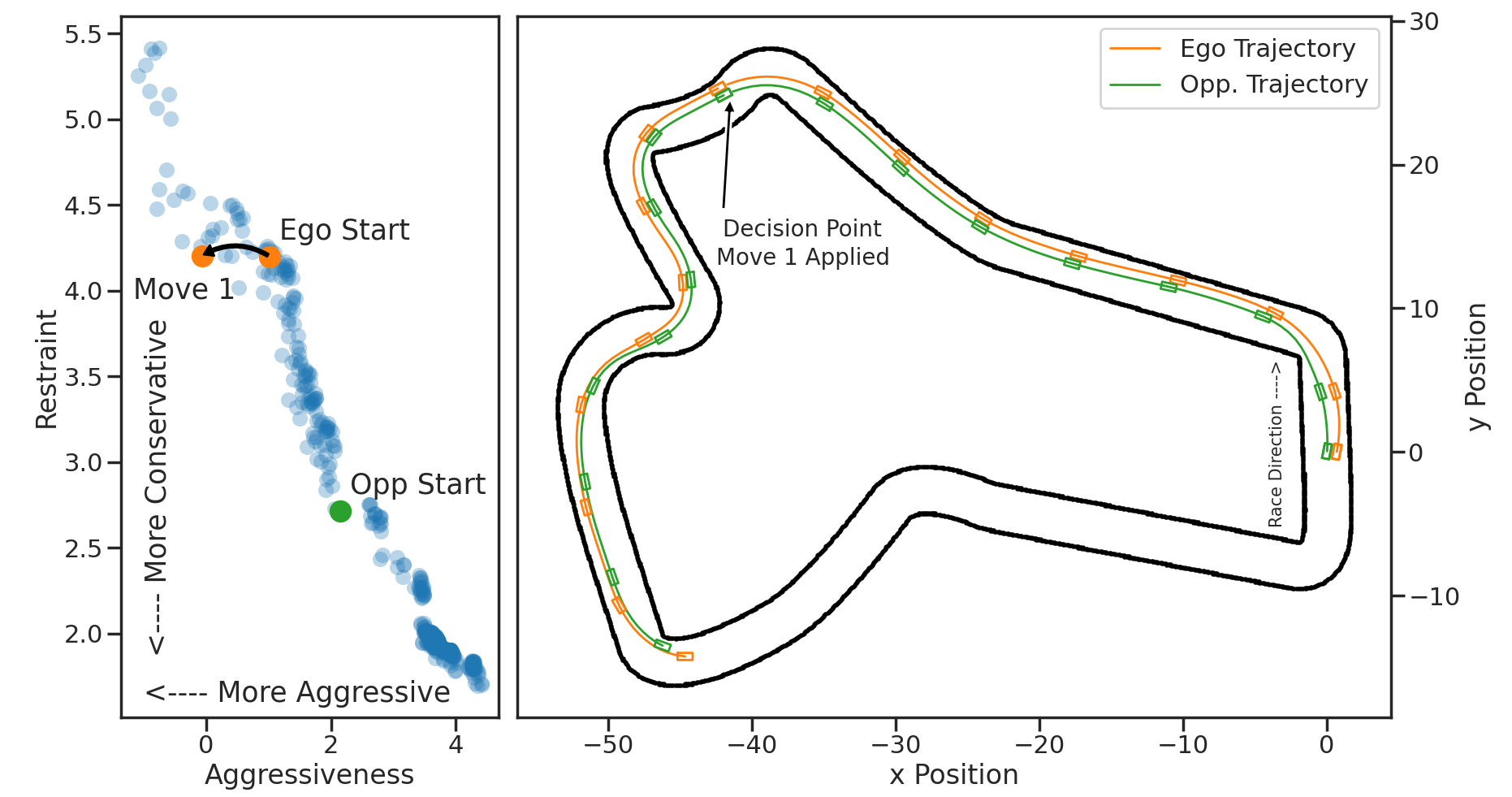}
    
    \caption{Effect of the strategy taking an action in the PCS. The left subplot shows what the action looks like in the Policy Characteristics Space. The right subplot shows that after taking an action, the ego overtakes the opponent.}
    \label{fig:make_one_move}
\end{figure}

To answer the second question, we examine selected rollouts of the racing games to investigate whether the actions of the agents are interpretable.
In the first segment of the track in Figure \ref{fig:make_one_move}, the ego agent observe the opponent for 8 seconds. The ego and the opponent remained side by side in the first segment. Then, at the decision point, the ego observed that the opponent's strategy produced a trajectory that indicates a more conservative than aggressive policy. The counterfactual regret approximator then predicted that increasing aggressiveness has the highest estimated counterfactual regret, and the strategy switches to a new policy (left subplot). The selected action's effect is immediately evident since the ego slowed down less than the opponent in the chicane and overtook the opponent by the end of the second step in the game.

\begin{figure}[h]
    \centering
    \includegraphics[trim={0cm 1cm 0cm 1cm},width=0.98\columnwidth]{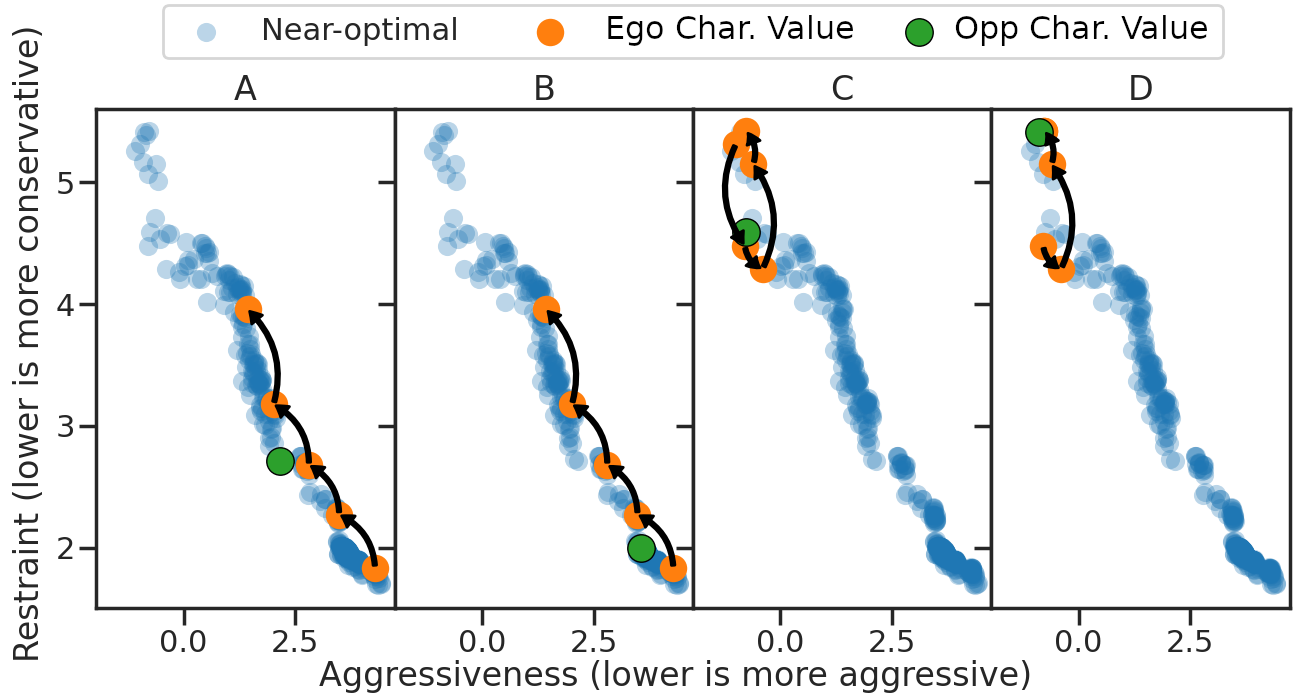}
    \caption{Trajectories of ego's actions in the PCS. In subplots A and B, the opponent is more conservative and the ego decides to increase aggressiveness right away. In subplots C and D, the opponent is more aggressive and the ego decides to increase in restraint until there is an opportunity to increase in aggressiveness and overtake.}
    \label{fig:moves}
    \vspace{-13pt}
\end{figure}

In Figure \ref{fig:moves}, four different rollouts are shown in which the ego wins at the end of the game.
In subfigures A and B, the opponent agent is observed to be in the lower right quadrant, meaning that these agents value safety more than progress. In these scenarios, the ego agent, starting in the lower right quadrant, became more aggressive than conservative with each action.
In subfigures C and D, the opponent agent is observed to be in the upper left quadrant, meaning that these agents value progress more than safety. In these scenarios, the ego decides to stay conservative and only increases aggressiveness later on when there is an opportunity to overtake.
From these two closer examinations, the use of interpretable policy characteristics functions has a clear benefit in explaining agent decisions.

\subsubsection{Encoding strategies in policy characteristics space preserves policy generalizability}
And lastly, to answer the third question, we examine the results when faced with unkown opponents in unknown environments.
Since we defined the policy characteristics functions to take the outcome (a set of state space trajectories) of a policy as input, agents do not need to assume a specific behavior model for the opponent, therefore, deal with unseen adversaries in unseen environments better.
The effect is evident in rows 3 and 6 in Table \ref{tab:online_exp}, we see that the increase in the win rate is retained even though facing unseen opponents during training (row 3), and in unseen environments (row 6).

\section{Limitation and Future Work}
The framework we propose can effectively bridge the gap between discrete agent strategies in game theory and continuous motion planning in dynamic environments, but there are still some limitations.
First, we chose to partition the continuous racing game into equal duration segments where strategies can perform actions. If agent decision-making is desired at a higher frequency, one might need to employ a receding-horizon update scheme. However, this might lead to too large a tree depth when training the regret approximator. Future work should focus on finding the balance point for this trade-off or new approaches to eliminate the need for partitioning. Second, instead of playing against randomized opponents during policy synthesis, the opponent set should become more and more competitive as the optimization iterates. We experimented with periodically mixing agents on the Pareto frontier into the opponent set, resulting in premature convergence to less competitive agents. Lastly, the policy characteristics functions $g_{\text{agg}}$ and $g_{\text{res}}$ used in the case study are chosen by hand with some domain knowledge. Future research can focus on discovering these functions automatically using disentanglement representation learning approaches to construct the PCS.

\section{Conclusion}
To bridge the gap between continuous motion planning and discrete strategies in a POSG setting, we propose a strategy representation with Policy Characteristic Space. This policy space formulation allows for interpretable discrete strategy actions while preserving continuous control.
We first define policies, strategies, and the Policy Characteristics Space.
Offline, we perform agent policy synthesis via multi-objective optimization and train a counterfactual regret approximator.
Online, we implement a planning pipeline that uses the approximated CFR to compete against an opponent.
In experiments, we provide statistical evidence showing significant improvements to the win rate that are generalized to unseen environments.
Lastly, we provide an examination on how the strategy representation improves interpretability when explaining agent decisions.

\def\UrlBreaks{\do\/\do-}
\bibliographystyle{ieeetr}
\bibliography{references, shuo_ref}

\begin{thebibliography}{10}

\bibitem{zanardi2021game}
A.~Zanardi, S.~Bolognani, A.~Censi, and E.~Frazzoli, ``Game theoretical motion planning: Tutorial icra 2021,'' 2021.

\bibitem{sobel_continuous_1973}
M.~J. Sobel, ``Continuous stochastic games,'' {\em Journal of Applied Probability}, vol.~10, pp.~597--604, Sept. 1973.
\newblock Publisher: Cambridge University Press.

\bibitem{schulman2017proximal}
J.~Schulman, F.~Wolski, P.~Dhariwal, A.~Radford, and O.~Klimov, ``Proximal policy optimization algorithms,'' {\em arXiv preprint arXiv:1707.06347}, 2017.

\bibitem{haarnoja2018soft}
T.~Haarnoja, A.~Zhou, P.~Abbeel, and S.~Levine, ``Soft actor-critic: Off-policy maximum entropy deep reinforcement learning with a stochastic actor,'' in {\em International conference on machine learning}, pp.~1861--1870, PMLR, 2018.

\bibitem{fridovich2020efficient}
D.~Fridovich-Keil, E.~Ratner, L.~Peters, A.~D. Dragan, and C.~J. Tomlin, ``Efficient iterative linear-quadratic approximations for nonlinear multi-player general-sum differential games,'' in {\em 2020 IEEE international conference on robotics and automation (ICRA)}, pp.~1475--1481, IEEE, 2020.

\bibitem{patil2023risk}
A.~Patil, Y.~Zhou, D.~Fridovich-Keil, and T.~Tanaka, ``Risk-minimizing two-player zero-sum stochastic differential game via path integral control,'' in {\em 2023 62nd IEEE Conference on Decision and Control (CDC)}, pp.~3095--3101, IEEE, 2023.

\bibitem{seyde_is_2021}
T.~Seyde, I.~Gilitschenski, W.~Schwarting, B.~Stellato, M.~Riedmiller, M.~Wulfmeier, and D.~Rus, ``Is {Bang}-{Bang} {Control} {All} {You} {Need}? {Solving} {Continuous} {Control} with {Bernoulli} {Policies},'' in {\em Advances in {Neural} {Information} {Processing} {Systems}}, vol.~34, pp.~27209--27221, Curran Associates, Inc., 2021.

\bibitem{sinha_formulazero_2020}
A.~Sinha, M.~O’Kelly, H.~Zheng, R.~Mangharam, J.~Duchi, and R.~Tedrake, ``{FormulaZero}: {Distributionally} robust online adaptation via offline population synthesis,'' in {\em International {Conference} on {Machine} {Learning}}, pp.~8992--9004, PMLR, 2020.

\bibitem{schwarting_stochastic_2021}
W.~Schwarting, A.~Pierson, S.~Karaman, and D.~Rus, ``Stochastic {Dynamic} {Games} in {Belief} {Space},'' {\em IEEE Transactions on Robotics}, vol.~37, pp.~2157--2172, Dec. 2021.

\bibitem{lanctot_unified_2017}
M.~Lanctot, V.~Zambaldi, A.~Gruslys, A.~Lazaridou, K.~Tuyls, J.~Perolat, D.~Silver, and T.~Graepel, ``A {Unified} {Game}-{Theoretic} {Approach} to {Multiagent} {Reinforcement} {Learning},'' in {\em Advances in {Neural} {Information} {Processing} {Systems}}, vol.~30, Curran Associates, Inc., 2017.

\bibitem{balduzzi2019open}
D.~Balduzzi, M.~Garnelo, Y.~Bachrach, W.~Czarnecki, J.~Perolat, M.~Jaderberg, and T.~Graepel, ``Open-ended learning in symmetric zero-sum games,'' in {\em International Conference on Machine Learning}, pp.~434--443, PMLR, 2019.

\bibitem{mcaleer2020pipeline}
S.~McAleer, J.~B. Lanier, R.~Fox, and P.~Baldi, ``Pipeline psro: A scalable approach for finding approximate nash equilibria in large games,'' {\em Advances in neural information processing systems}, vol.~33, pp.~20238--20248, 2020.

\bibitem{liu2022neupl}
S.~Liu, L.~Marris, D.~Hennes, J.~Merel, N.~Heess, and T.~Graepel, ``Neupl: Neural population learning,'' {\em arXiv preprint arXiv:2202.07415}, 2022.

\bibitem{hafner_dream_2020}
D.~Hafner, T.~Lillicrap, J.~Ba, and M.~Norouzi, ``Dream to {Control}: {Learning} {Behaviors} by {Latent} {Imagination},'' Mar. 2020.
\newblock arXiv:1912.01603 [cs].

\bibitem{schwarting_deep_2021}
W.~Schwarting, T.~Seyde, I.~Gilitschenski, L.~Liebenwein, R.~Sander, S.~Karaman, and D.~Rus, ``Deep {Latent} {Competition}: {Learning} to {Race} {Using} {Visual} {Control} {Policies} in {Latent} {Space},'' Feb. 2021.
\newblock arXiv:2102.09812 [cs].

\bibitem{xie_learning_2021}
A.~Xie, D.~Losey, R.~Tolsma, C.~Finn, and D.~Sadigh, ``Learning {Latent} {Representations} to {Influence} {Multi}-{Agent} {Interaction},'' in {\em Proceedings of the 2020 {Conference} on {Robot} {Learning}}, pp.~575--588, PMLR, Oct. 2021.
\newblock ISSN: 2640-3498.

\bibitem{mnih_playing_2013}
V.~Mnih, K.~Kavukcuoglu, D.~Silver, A.~Graves, I.~Antonoglou, D.~Wierstra, and M.~Riedmiller, ``Playing {Atari} with {Deep} {Reinforcement} {Learning},'' Dec. 2013.
\newblock arXiv:1312.5602 [cs].

\bibitem{jin_regret_2018}
P.~Jin, K.~Keutzer, and S.~Levine, ``Regret {Minimization} for {Partially} {Observable} {Deep} {Reinforcement} {Learning},'' Oct. 2018.
\newblock arXiv:1710.11424 [cs].

\bibitem{brown_deep_2018}
N.~Brown, A.~Lerer, S.~Gross, and T.~Sandholm, ``Deep {Counterfactual} {Regret} {Minimization},'' Nov. 2018.

\bibitem{hansen_dynamic_2004}
E.~A. Hansen, D.~S. Bernstein, and S.~Zilberstein, ``Dynamic {Programming} for {Partially} {Observable} {Stochastic} {Games},'' 2004.

\bibitem{hansen_cma_2016}
N.~Hansen, ``The {CMA} {Evolution} {Strategy}: {A} {Tutorial},'' Apr. 2016.

\bibitem{fonseca_improved_2006}
C.~Fonseca, L.~Paquete, and M.~Lopez-Ibanez, ``An {Improved} {Dimension}-{Sweep} {Algorithm} for the {Hypervolume} {Indicator},'' in {\em 2006 {IEEE} {International} {Conference} on {Evolutionary} {Computation}}, pp.~1157--1163, July 2006.
\newblock ISSN: 1941-0026.

\bibitem{zinkevich_regret_2007}
M.~Zinkevich, M.~Johanson, M.~Bowling, and C.~Piccione, ``Regret {Minimization} in {Games} with {Incomplete} {Information},'' in {\em Advances in {Neural} {Information} {Processing} {Systems}}, vol.~20, Curran Associates, Inc., 2007.

\bibitem{okelly_f1tenth_2020}
M.~O'Kelly, H.~Zheng, D.~Karthik, and R.~Mangharam, ``{F1TENTH}: {An} {Open}-source {Evaluation} {Environment} for {Continuous} {Control} and {Reinforcement} {Learning},'' {\em Proceedings of Machine Learning Research}, vol.~123, Apr. 2020.

\bibitem{brockman_openai_2016}
G.~Brockman, V.~Cheung, L.~Pettersson, J.~Schneider, J.~Schulman, J.~Tang, and W.~Zaremba, ``{OpenAI} {Gym},'' June 2016.
\newblock arXiv:1606.01540 [cs].

\bibitem{althoff_commonroad_2017}
M.~Althoff, M.~Koschi, and S.~Manzinger, ``{CommonRoad}: {Composable} benchmarks for motion planning on roads,'' in {\em 2017 {IEEE} {Intelligent} {Vehicles} {Symposium} ({IV})}, pp.~719--726, IEEE, 2017.

\bibitem{kelly_reactive_2003}
A.~Kelly and B.~Nagy, ``Reactive nonholonomic trajectory generation via parametric optimal control,'' {\em The International Journal of Robotics Research}, vol.~22, no.~7-8, pp.~583--601, 2003.
\newblock Publisher: SAGE Publications.

\bibitem{noauthor_implementation_nodate}
``Implementation of the {Pure} {Pursuit} {Path} {Tracking} {Algorithm},'' tech. rep.
\newblock Section: Technical Reports.

\bibitem{f1tenth_f1tenth_nodate}
f1tenth, ``{F1TENTH} - {ICRA} 2022 {Competition} {Results}.''

\bibitem{zhijunzhuang_zzjun725f1tenth-racing-stack-icra22_2024}
ZhijunZhuang, ``zzjun725/f1tenth-racing-stack-{ICRA22},'' Jan. 2024.
\newblock original-date: 2023-10-02T13:57:49Z.

\bibitem{kulesza_determinantal_2012}
A.~Kulesza and B.~Taskar, ``Determinantal point processes for machine learning,'' {\em Foundations and Trends® in Machine Learning}, vol.~5, no.~2-3, pp.~123--286, 2012.
\newblock arXiv:1207.6083 [cs, stat].

\bibitem{schulman_proximal_2017}
J.~Schulman, F.~Wolski, P.~Dhariwal, A.~Radford, and O.~Klimov, ``Proximal {Policy} {Optimization} {Algorithms},'' Aug. 2017.
\newblock arXiv:1707.06347 [cs].

\bibitem{huang_cleanrl_2022}
S.~Huang, R.~F.~J. Dossa, C.~Ye, J.~Braga, D.~Chakraborty, K.~Mehta, and J.~G.~M. Araújo, ``{CleanRL}: {High}-quality {Single}-file {Implementations} of {Deep} {Reinforcement} {Learning} {Algorithms},'' {\em Journal of Machine Learning Research}, vol.~23, no.~274, pp.~1--18, 2022.

\end{thebibliography}


\end{document}